\documentclass[sn-mathphys,Numbered]{sn-jnl}

\usepackage{graphicx}%
\usepackage{multirow}%
\usepackage{amsmath,amssymb,amsfonts}%
\usepackage{amsthm}%
\usepackage{mathrsfs}%
\usepackage[title]{appendix}%
\usepackage{xcolor}%
\usepackage{textcomp}%
\usepackage{manyfoot}%
\usepackage{booktabs}%
\usepackage{algorithm}%
\usepackage{algorithmicx}%
\usepackage{algpseudocode}%
\usepackage{listings}%

\theoremstyle{thmstyleone}%
\theoremstyle{thmstyletwo}%

\theoremstyle{thmstylethree}%

\begin{document}

\title[Article Title]{Joint Self-supervised Depth and Optical Flow Estimation towards Dynamic Objects}


\author[1]{\fnm{Zhengyang} \sur{Lu}}\email{7191905018@stu.jiangnan.edu.cn}

\author*[1]{\fnm{Ying} \sur{Chen}}\email{chenying@jiangnan.edu.cn}

\affil[1]{\orgdiv{the Key Laboratory of Advanced Process Control for Light Industry (Ministry of Education)}, \orgname{Jiangnan University}, \orgaddress{\city{Wuxi}, \postcode{214026}, \country{China}}}


\abstract{Significant attention has been attracted to deep learning-based depth estimates.
	Dynamic objects become the most hard problems in inter-frame-supervised depth estimates due to the uncertainty in adjacent frames.
	Thus, integrating optical flow information with depth estimation is a feasible solution, as the optical flow is an essential motion representation.
	In this work, we construct a joint inter-frame-supervised depth and optical flow estimation framework, which predicts depths in various motions by minimizing pixel wrap errors in bilateral photometric re-projections and optical vectors.
	For motion segmentation, we adaptively segment the preliminary estimated optical flow map with large areas of connectivity.
	In self-supervised depth estimation, different motion regions are predicted independently and then composite into a complete depth. 
	Further, the pose and depth estimations re-synthesize the optical flow maps, serving to compute reconstruction errors with the preliminary predictions.
	Our proposed joint depth and optical flow estimation outperforms existing depth estimators on the KITTI Depth dataset, both with and without Cityscapes pretraining. Additionally, our optical flow results demonstrate competitive performance on the KITTI Flow 2015 dataset.}




\maketitle

\section{Introduction}
With the explosion of deep-learning technologies, depth estimation demonstrates promise for stereoscopic perception in complex scenes, which facilitates high-level computer visions, involving human-machine understanding\cite{yu2019hierarchical,hong2015multimodal}, stereoscopic perception, scene segmentation, driving assistance and behaviour prediction\cite{hong2018multimodal}. 
In fact, bio-vision systems can perceive real-world scenes without barriers, whereby systems pre-trained with sufficient prior information can measure accurate depth maps. 
While the binocular mechanism is widely spread in bio-vision systems, depth perception remains sensitive in monocular conditions. 
Besides, inferring depths from single images with deep-learning models remain exceedingly challenging, as an ill-posed vision task.

Deep learning-based depth estimators have been extensively explored for years, yielding unparalleled accuracies against classic methods.
Existing supervised models\cite{laina2016deeper,fu2018deep,liu2015learning,zhang2018progressive,li2015depth,lu2021ga} can predict accurate depths from monocular images by formulating the depth estimates as a regression issue.
Godard\cite{godard2017unsupervised} provided a consistent binocular framework, allowing supervision by left-right pairs without labelled depths.
Lu\cite{lu2022pyramid} leveraged the Fourier perspective to construct a robust depth estimator with a pyramid frequency network.
The Mono-Former\cite{bae2022monoformer}, the first CNN-Transformer for depth estimation, was conceived for multi-scene generalization.  

Self-supervised depth estimators provide a universal framework with binocular stereo images or continuous frame supervision, which alleviates laborious annotation works\cite{xiang2021self, wei2022triaxial}.
The inter-frame-supervised method was first proposed as Monodepth2\cite{godard2019digging}, providing a label-free depth framework via joint estimation of camera poses and inverse depths.
Johnston\cite{johnston2020self} leveraged a self-attentive mechanism and discrete disparity reconstruction to learn accurate depths in self-supervision.
Guizilini\cite{guizilini2022learning} presented a multi-task framework, simultaneously estimating depth, optical flow, and scene flow to integrate multiple tasks via image synthesis and geometric constraints.
Recurrent Multi-Scale Feature Modulation(RMSFM)\cite{zhou2021r} designed multi-scale modulations with successive depth updates to improve the coarse-to-fine performance.
Due to the neglect of contextual consistency between multi-scale features, Guizilini\cite{guizilini2020semantically} introduced the Self-Distilled Feature Aggregation (SDFA) module, which enables simultaneous aggregation of low-scale and high-scale features while maintaining contextual consistency.

Two common solutions to the problem of dynamic objects in depth estimation methods, which incorporate optical flow information, can be found in the literature. The first solution involves using optical flow to track the motion of dynamic objects and refine the depth map\cite{nekrasov2019real}. The second solution leverages information from motion segmentation to identify dynamic objects and remove the impact of their dynamic characteristics from the depth map\cite{mousavian2016joint}.

In stationary scenes with moving viewpoints, the optical flow map carries the same information as the camera transformation and the depth map from the inter-frame-supervised methods. 
In other words, ideal optical flow maps can be equivalently decomposed into camera transformations and depths without occlusion components.
Hence, camera pose estimation in inter-frame supervision can be considered as a regression issue, estimating eigenvalues from static components that dominate the scene.

In order to construct a collaborative framework that focuses on dynamic objects, we unite two intrinsically homogeneous tasks, namely inter-frame-supervised depth and optical flow estimation.
First, independent motion direction regions are separated from the optical flow estimation results.
Next, each segmented region is fed into the depth module to predict inverse depths and camera transformation, respectively.
In addition, the optical flow, depth and pose network are constrained by bilateral photometric re-projection loss and optical flow reconstruction loss, which are derived from the estimated depths and camera transformation.
Relative to established self-supervised depth estimation approaches, the novel method exhibits remarkable improvements in accuracy, attributable to the advancements in addressing dynamic object problem. Simultaneously, the bidirectional reprojection constraint bolsters the robustness of the self-supervised mechanism.
Specifically, the multi-task framework focusing on dynamic objects outperforms existing researches on the KITTI Depth dataset.

The contributions of the multi-task framework are outlined:
\begin{itemize}
	
	\item{We construct a joint inter-frame-supervised depth and optical flow estimation framework, which predicts depths in different motions by minimizing pixel wrap errors between the photometric re-projections and optical vectors.}
	\item{In optical flow-based motion segmentation, we adaptively segment the preliminary estimated optical flow map by connectivity.} 
	\item{For bilateral inter-frame-supervised depth estimates, each motion region is predicted independently before the complete depth map composition. 
		Further, the pose and depth predictions re-synthesize the optical flow maps, serving to compute synthesis errors with preliminary predictions.} 
	\item{The proposed joint framework outperforms advanced depth and optical flow estimators on KITTI Depth and Flow dataset.}
\end{itemize}

\begin{figure*}[htbp]
	\centering
	\includegraphics[width=0.98\linewidth]{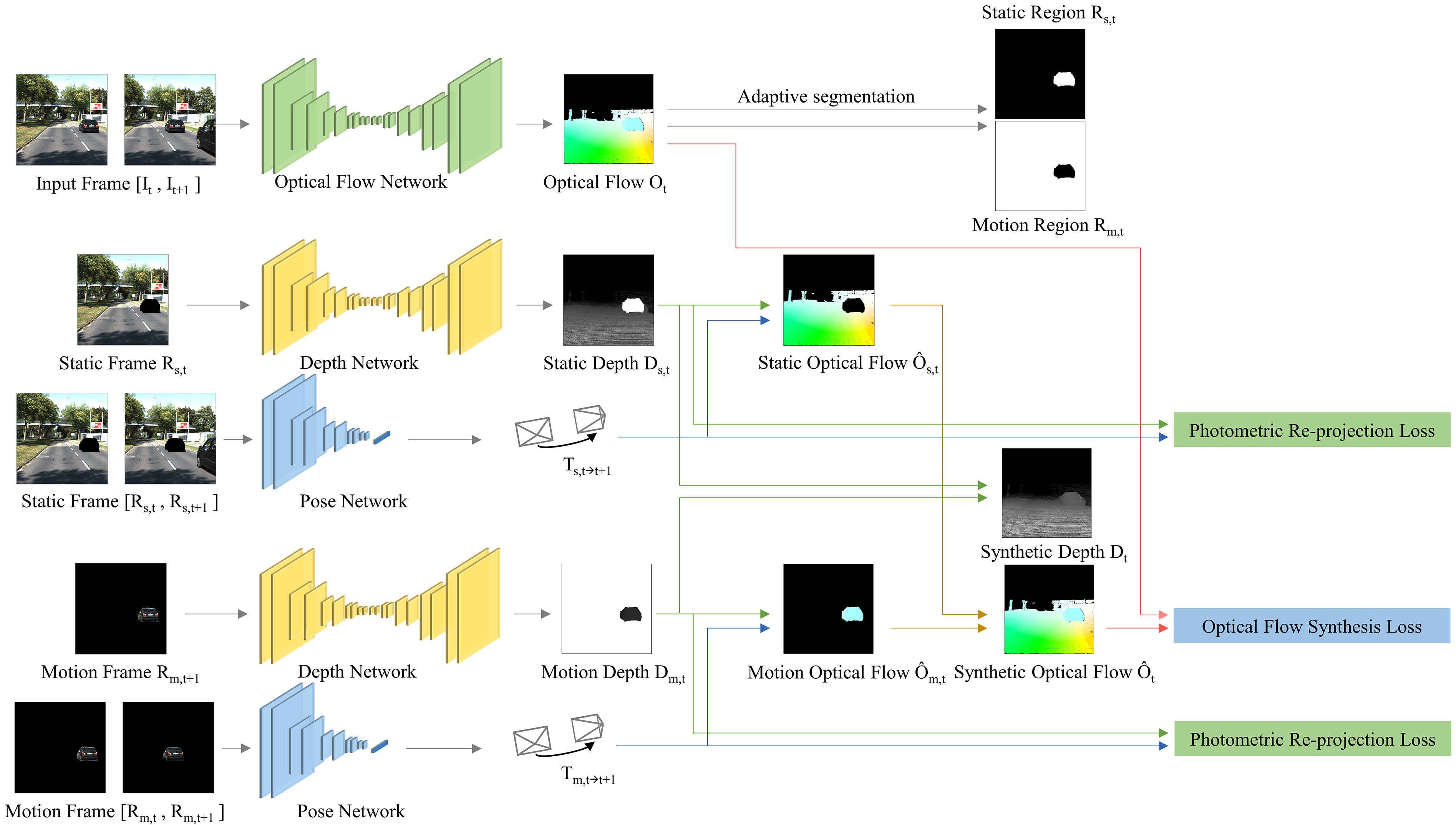}
	\caption{The overview of the joint optical flow and inter-frame-supervised depth estimation towards dynamic objects. Depth networks for static and motion components share the same weights, as do the pose networks.}
	\label{fig:detailarc}
\end{figure*}

\section{Methodology}

To constrain the optical flow and ego-motion consistency, we demonstrate an inter-frame-supervised depth and optical flow estimation framework, which predicts depths by minimizing pixel wrap errors between the photometric re-projections and optical vectors.

\subsection{Overview}

As indicated in Fig.\ref{fig:detailarc}, the joint depth and optical framework focusing on the dynamic objects comprises three modules: 1) Optical flow-based motion segmentation; 2) Bilateral inter-frame-supervised depth estimation; and 3) Optical flow synthesis.
The optical flow-based motion segmentation is intended to separate pixel regions with heterogeneous motion directions.
Then, depth and pose estimations are performed independently in dynamic and static regions to compute re-projection errors with bilateral constraints.
Finally, the optical flow map can be reconstructed from the predicted depths and camera pose, whose endpoint errors with the raw optical flow optimize the two-stage framework.

Optical flow-based motion segmentation serves a critical function in the network. The purpose of this module is to distinguish between pixel regions that exhibit heterogeneous motion directions. Optical flow, essentially the pattern of apparent motion of objects, surfaces, and edges in a visual scene caused by the relative motion between an observer and the scene, is used to effectively segment the image into regions based on the direction and magnitude of motion. This segmentation process allows the network to handle complex scenes where multiple objects may be moving in different directions.

Following this segmentation process, depth and pose estimations are conducted independently in both dynamic and static regions. The aim here is to compute re-projection errors with bilateral constraints. The depth estimation is performed using a bilateral inter-frame-supervised approach, which takes into account both the previous and subsequent frames to make more accurate depth estimations. The pose estimation, on the other hand, is concerned with determining the orientation and position of the camera relative to the scene. The bilateral constraints act as a regulatory mechanism to ensure that these estimations remain consistent and accurate across all frames.

Lastly, the optical flow map is reconstructed from the predicted depths and the estimated camera pose. This reconstructed optical flow map provides a detailed representation of the motion within the scene. The endpoint errors, which are the differences between the reconstructed optical flow map and the original optical flow, are then used to optimize the two-stage framework. This process is instrumental in refining the performance of the system, allowing it to improve its accuracy over time and adapt to changing conditions.

\subsection{Optical flow-based motion segmentation}

Following FlowNet~\cite{dosovitskiy2015flownet}, a standard U-net~\cite{ronneberger2015u} is leveraged to predict preliminary optical flow maps, which guides the motion separation.

In adjacent frames, ideal perspective-variable regions provide continuous optical vectors.
Rigid object in relative motion is considered as virtual perspective transformations, that is, relative motion regions represent continuous vectors.
Therefore, it is feasible to segment relative moving components in the same scene with the optical flow method.

To segment regions with heterogeneous motion direction, the preliminary predicted optical flow requires mean convolution operations to smooth the vectors due to crude output.
To retrieve sharp outlines, a Sobel operator is applied to filter the smoothed optical flow map.
Finally, the main relative motion regions are selected by filling the approximately enclosed outline according to the given boundary threshold. These regions are determined by an eight-connected pixel traversal~\cite{haralick1992computer}.
For further processing, segmentation areas are padded with zero pixel values.
Furthermore, if massive motion components are erroneously segmented as a static region, their pose estimation is unique. In other words, only the dominant camera transformations are obtained in wrong segmentations and motion forms of small misplaced regions are omitted.
Hence, the error in the inter-frame-supervised depth module arises from the pixel sets whose motion forms are erroneously represented.
It is worth noting that these region segmentation errors are penalized in the optical flow reconstruction loss.

\subsection{Bilateral Inter-frame-supervised depth estimation}

As results from optical flow-based segmentation, components with heterogeneous motion directions are separated.
We prefer to address static components as primary motion direction regions and dynamic ones as minor regions, as motion is absolute in essence.
For the primary motion direction regions, a VGG-based PoseNet~\cite{simonyan2014very} is applied to estimate the ego-motion between adjacent static frames $R_{s,t}$ and $R_{s,t+1}$:
\begin{equation}
	\begin{split}
		T_{s,t\rightarrow t+1}&=PoseNet\left(R_{s,t}, R_{s,t+1}\right)\\
		R_{s,t\rightarrow {t+1}} &= R_{s,t}\langle project(D_{s,t}, T_{s,t+1\rightarrow t}, K)\rangle\\	
	\end{split}
\end{equation}

Besides, the corresponding backward re-projection process can be expressed as:
\begin{equation}
	\begin{split}
		T_{s,t+1\rightarrow t}&=PoseNet\left(R_{s,t+1}, R_{s,t}\right)\\
		R_{s,t+1\rightarrow t} &= R_{s,t+1}\langle project(D_{s,t+1}, T_{s,t\rightarrow t+1}, K)\rangle		
	\end{split}
\end{equation}
where $T$ donates the camera pose transformation between two frames, $K$ donates the camera intrinsic parameters, $\langle \rangle$ donates the per-pixel sampling~\cite{stn} and $project()$ donates the coordinate re-projection~\cite{zhou2017}.
Similar to primary motion regions, the forward and backward photometric re-projections for minor motion regions $R_{m,t\rightarrow {t+1}}$ and $R_{m,t+1\rightarrow t}$ are derived in the same way.
Therefore, the photometric error $L_{pe}$ comprises SmoothL1 and SSIM\cite{wang2004image}:
\begin{equation}
	L_{pe}(I_1, I_2) = \alpha(1 - \mathrm{SSIM} (I_1, I_2)) + (1 - 2\alpha)\| I_1 - I_2  \| _1.
\end{equation}
where $\alpha=0.45$. 

Following previous inter-frame-supervision works~\cite{godard2019digging}, to address scene occlusions, the bilateral photometric re-projection loss $\mathcal{L}_{ph,s}$ is deployed to the primary motion regions:
\begin{equation}
	\mathcal{L}_{ph,s} = L_{pe}(R_{s,t+1}, R_{s,t\rightarrow t+1})+L_{pe}(R_{s,t}, R_{s,t+1\rightarrow t})
\end{equation}

Same as $\mathcal{L}_{ph,s}$, the photometric re-projection loss $\mathcal{L}_{ph,m}$ for minor motion regions can be derived in similar operation.
Finally, by combining the various motion components, the integral re-projection loss $\mathcal{L}_{ph}$ is:
\begin{equation}
	\mathcal{L}_{ph} = \mathcal{L}_{ph,s}+\mathcal{L}_{ph,m}
\end{equation}

Above derivations only consider the two motion regions case, but real-world scenarios exist multi-motion regions, for example, multiple driving cars in lanes. 
Hence, the same re-projection method is adapted to count each heterogeneous motion individually, as an additional $\mathcal{L}_{ph,m}$. Therefore, the re-projection loss $L_{ph}$ for multiple motion regions is expressed as:

\begin{equation}
	L_{p h}=L_{p h, s}+\sum_{m=1}^{k} L_{p h, m}
\end{equation}
where $k$ donates the number of dynamic regions and $m$ donates the number of motion regions.

\subsection{Optical flow synthesis}

The optical flow is a composite pixel-level representation of the depth map and the camera transformation that allow interconversion in the static scene.
Thus, the reconstructed optical flow for static components $\hat{O}_{s,t}$ can be defined as:
\begin{equation}
	\hat{O}_{s,t}=project(D_{s,t}, T_{s,t+1\rightarrow t}, K)
\end{equation}

Obviously, motion components' optical flow $\hat{O}_{m,t}$ have the same form.
Then, we combine the static and motion components as:
\begin{equation}
	\hat{O}_{t}=\hat{O}_{s,t}+\hat{O}_{m,t}
\end{equation}

Following previous works, the optical flow module applies the endpoint error, which is the L2 distance between the vectors and predictions.
Hence, the optical flow synthesis loss $L_{flow}$ can be computed as:
\begin{equation}
	\mathcal{L}_{flow} =  \| \hat{O}_{t}-O_{t}  \| _2
\end{equation}

In network optimization, the depth network loss function applies re-projection loss and optical flow synthesis loss, while the pose network loss function also applies re-projection loss and flow synthesis loss for joint optimization. Therefore, the loss function for depth network $\mathcal{L}_{depth}$ and pose network $\mathcal{L}_{pose}$ can be formulated as:
\begin{equation}
	\begin{split}
		\mathcal{L}_{depth} &= \mathcal{L}_{ph}+\lambda\mathcal{L}_{flow}\\
		\mathcal{L}_{pose} &= \mathcal{L}_{ph}+\lambda\mathcal{L}_{flow} \\
	\end{split}
\end{equation}
where $\mathcal{L}_{ph}$ represents the re-projection loss, $\mathcal{L}_{flow}$ represents the flow synthesis loss, and $\lambda$ is the weight coefficient for the loss balance. The $\lambda$ is set to 0.1 based on the experimental results.

Meanwhile, the optical flow network loss function is optimized solely using flow reconstruction loss. The optical flow network loss function $\mathcal{L}_{optical}$ can be expressed as:

\begin{equation}
	\begin{split}
		\mathcal{L}_{optical} &= \mathcal{L}_{flow} \\
	\end{split}
\end{equation}

\section{Experiments}

\subsection{Experiments Settings}

\subsubsection{Datasets}
The KITTI depth prediction dataset~\cite{Geiger2013IJRR} is extensively employed for outdoor scene depth estimation, comprising 42,949 training, 1,000 validation and 500 test samples, which have sparse depth pixel annotations.
For network processing, images are scaled to 352$\times$1216 to adapt the convolution interface.
Median scaling~\cite{zhou2017} is implemented to normal scale values due to previous depth estimators being infeasible to capture certain scales.

\subsubsection{Metrics}

For fairness, relative depths are bounded to a given distance between 0$m$ and 120$m$ and compared with existing depth estimators by standard metrics: Absolute Relative Error (AbsRel), Square Relative Error (SqRel), Root Mean Square Error (RMS), Root Mean Square Error in Logarithmic operation(RMS(${log}$))~\cite{karsch2014depth} and Accuracies of three thresholds.

\subsubsection{Experiment Details}

The proposed framework is implemented on the PyTorch~\cite{paszke2019pytorch} platform and executed on 2 Nvidia RTX2080 GPUs.
We employ VGG-16~\cite{simonyan2014very} as the PoseNet encoder whose initial network adopts the pre-trained model's weights on ImageNet classification~\cite{deng2009imagenet}.
For the optical flow and depth network, standard end-to-end U-net backbones are deployed, which facilitates further deployment.
Furthermore, the learning rate for PoseNet is $10^{-4}$ and for depth and optical flow network are $10^{-3}$, which reduces into $10\%$ every 20 epochs.
In the motion segmentation, the smooth operation conducts three times with kernels of 3, 5 and 9, followed by a Sobel operation with a threshold of 0.5 and a motion area filter with a minimum of 3000 pixels.

\subsection{Ablation Experiments}

To determine hyper-parameters for motion segmentations, ablation experiments with various thresholds are exhibited in Table.\ref{tab:result_abl}. 'S' represents the Segmentation-based Method for depth estimation, which is based on optical flow segmentation, M represents the Monodepth2\cite{godard2019digging} and 'BiE' denotes the Bilateral Re-projection Error.

\begin{table}[tbp]
	\caption{Quantitative results with multiple settings, bilateral re-projection error(BiE) and minimum area loss $T_{R}$, on KITTI depth dataset.}
	\label{tab:result_abl}
	\begin{tabular}{r|c|cccc|ccccc}
		\toprule
		\specialrule{0em}{0pt}{2pt}	
		Method      	&   $T_{R}$        & AbsRel & SqRel & RMS   & RMS(${log}$) & $\delta_1$ & $\delta_2$ & $\delta_3$ \\ 
		\specialrule{0em}{0pt}{0pt}
		\midrule
		\specialrule{0em}{0pt}{2pt}	
		M(w/o  BiE)     &     -                        & 0.1150	& 	0.9030 	& 	4.8630	& 	0.1930	& 	0.877	& 	0.959	& 	0.981\\	
		S(w/o  BiE)     &     1000                     & 0.1050	&	0.7820 	&	4.5980 	&	0.1810 	&	0.886 	&	0.967 	&	0.984	\\
		S(w/o  BiE)     &     3000                     & 0.0970 & 	0.6470	&	3.9910	&	0.1690	&	0.899	&	0.968	&	0.984	\\
		S(w/o  BiE)     &     5000                     & 0.0980	&	0.6450	&	3.9980	&	0.1670	&	0.901	&	0.970	&	0.988	\\ \hline
		\specialrule{0em}{0pt}{1pt}	
		M(with BiE)     &     -                        & 0.1120	&	0.7880 	& 	4.6020 	& 	0.1900 	& 	0.873 	& 	0.961 	& 	0.981\\	
		S(with BiE)     &     1000                     & 0.1020	&	0.6900	&	4.2180	&	0.1701	&	0.898	&	0.969	&	0.987	\\
		S(with BiE)  	&     3000                     & 0.0950 & 	0.6180 	& 	3.9400	& 	0.1680 	&  	0.904 	& 	0.969 	& 	0.988\\
		S(with BiE)   	&     5000                     & 0.0960	&	0.6390	&	3.9720	&	0.1689	&	0.900	&	0.968	&	0.985	\\ 
		\specialrule{0em}{0pt}{0pt}	
		\bottomrule
	\end{tabular}
\end{table}

\begin{figure}[h]
	\centering
	\includegraphics[width=\linewidth]{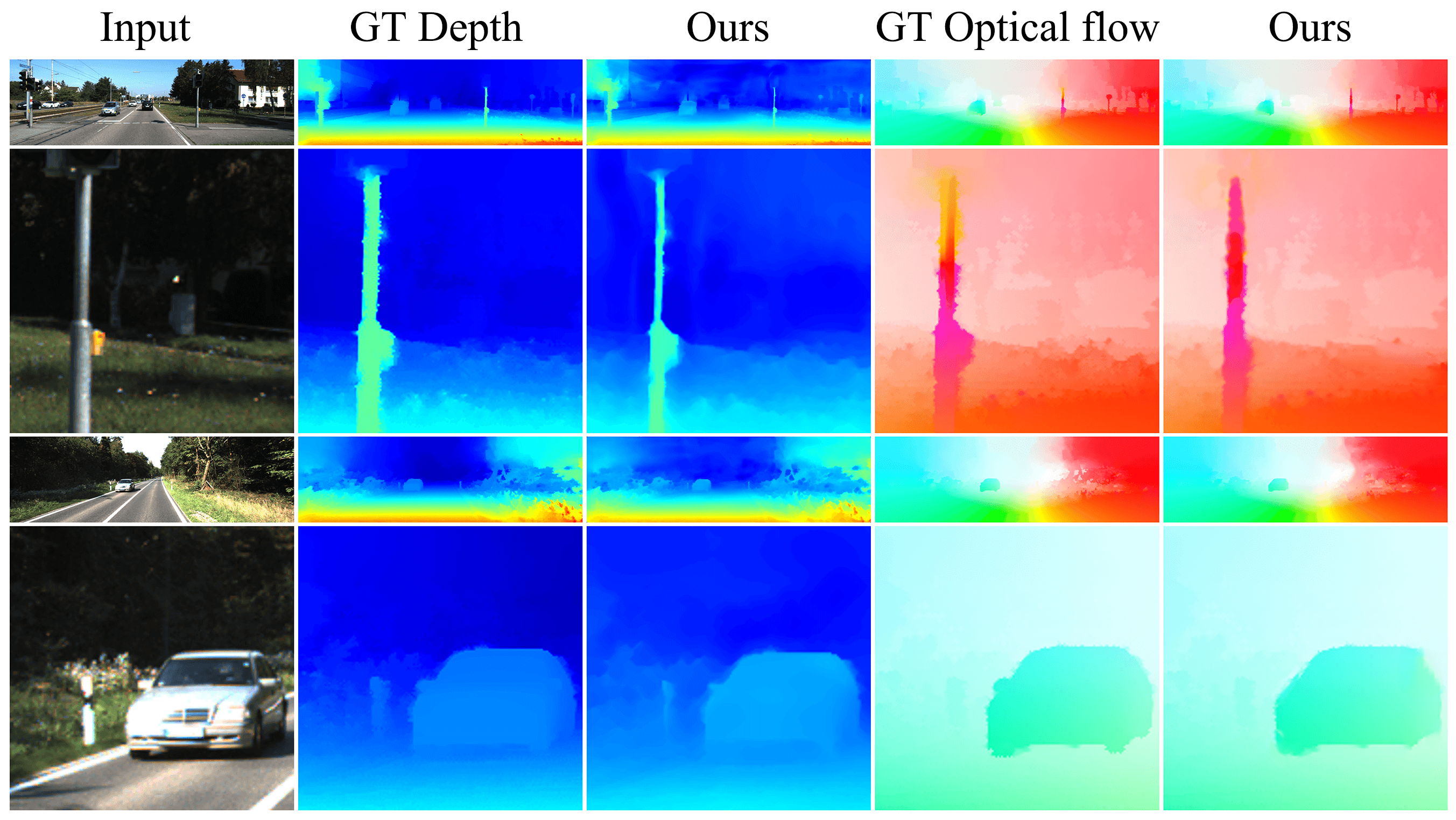}
	\caption{Visual results and zoomed objects on the Eigen splits. Both depth and optical flow results are provided for comparisons with ground-truths.}
	\label{fig:ablation}
\end{figure}

As expected, the bilateral constraint substantially improves each motion region's pose and depth estimation.
Meanwhile, the minimum area setting filters the small and incorrect motion regions.
Among the above operations, the optical flow-based motion segmentation provides crucial improvements, achieving 0.0950 error on AbsRel, 0.6180 on SqRel, 3.940 on RMS and 0.1680 on RMS(${log}$).
Compared to the original monodepth2, the most crucial enhancement in the proposed methodology is attributed to the dynamic object segmentation mechanism, which results in an 8.9\% decrease in AbsRel. Concurrently, the bidirectional constraint contributes to a significant improvement, approximately a 2.8\% decrease in AbsRel.
In the multivariate experiments, the model selected the optimal combination, which corresponds to a minimum area filter value of 3,000 pixels and a bilateral error constraint. 
The ablation experiments reveal that each threshold remarkably improved all 7 metrics.
Among the thresholds, the bilateral constraint brings the most potent improvement, which means that most noise in the optical flow map is successfully filtered.
Moreover, visual depths and optical flows maps are exemplified in Fig.\ref{fig:ablation}.

As illustrated in Fig.\ref{fig:ablation}, the ablation experiments with the optimal combination showcase accurate visual depth maps and optical flow maps, which are visually consistent with the ground truth depth results. Specifically, the proposed method successfully reconstructs the slender lampposts, although there is an inconsistency in the thickness of the lampposts' upper and lower ends. Compared to the ground-truth optical flow map, the lamppost optical flow estimated by the proposed method appears visually more reasonable.

\subsection{Depth Comparison with Existing Methods}

In this section, we conduct a quantitative and qualitative comparison of existing depth estimation methods on the KITTI dataset. The experimental results analyze the performance of various depth estimation techniques based on inter-frame supervision mechanisms, which include multiple non-pretrained self-supervised depth estimation methods.

\begin{figure}[h]
	\centering
	\includegraphics[width=\linewidth]{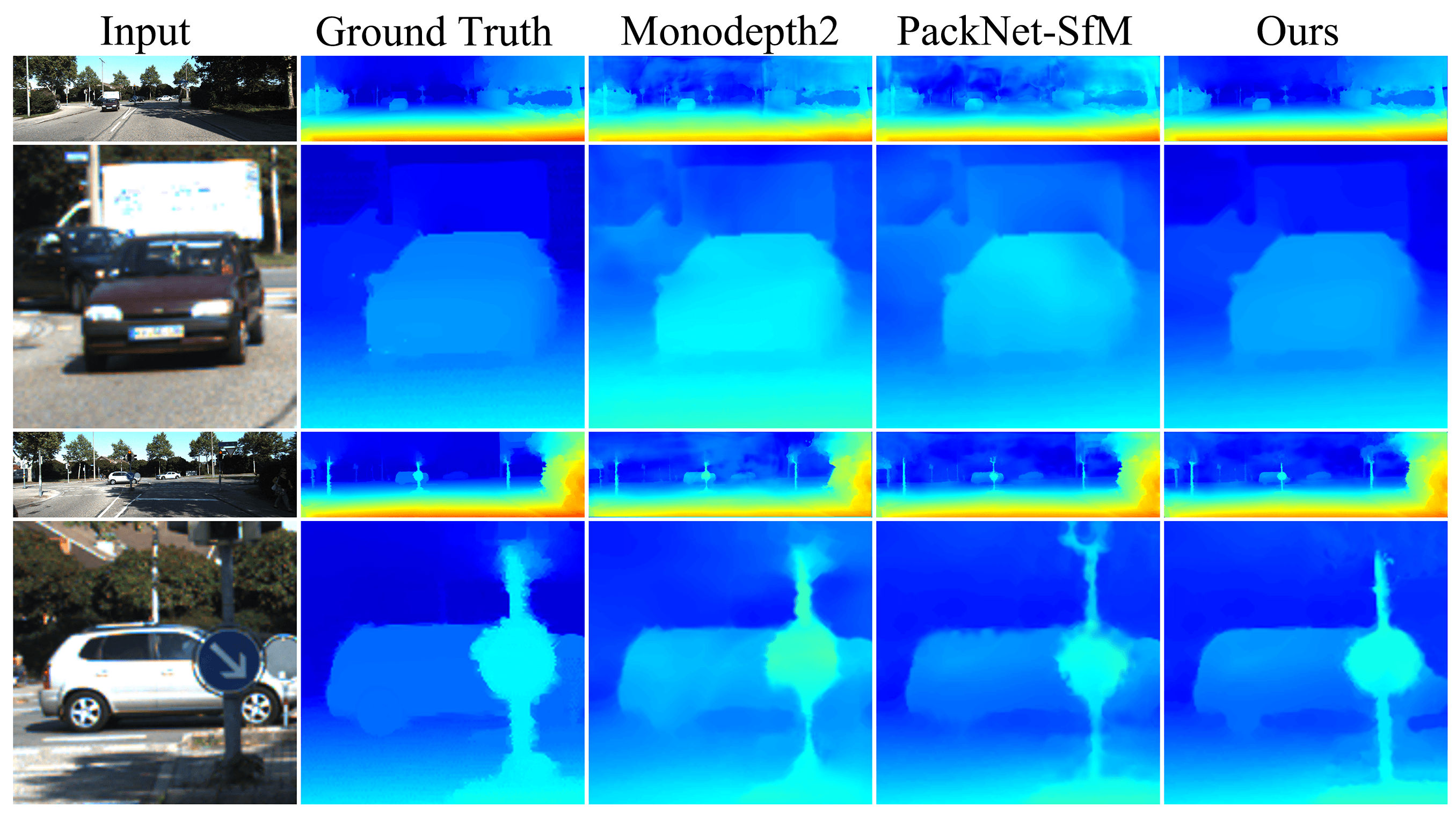}
	\caption{Visual results on KITTI with inter-frame-supervised methods. Our method is superior to advanced methods in edge sharpness for occluded objects.}
	\label{fig:quantitive}
\end{figure}

In Fig.\ref{fig:quantitive}, Most existing methods successfully estimate the lane scene's depth maps. Among these methods, the proposed method with the joint depth and optical flow estimation framework significantly outperforms existing methods, particularly in predicting the occluded areas of objects, as seen in the car edges and lamppost reconstruction in the upper and lower images, respectively. The primary reason for this performance improvement is that the optical flow estimation can approximate the relative position relationships between occlusions and the scene, thus assisting in depth prediction.

\begin{table}[t]
	\caption{Quantitative depth results on Eigen split. Extensive depth estimators are trained on KITTI depth(K).}
	\label{tab:result_kitti}
		\begin{tabular}{l|c|cccc|ccccccc}
			\toprule
			\specialrule{0em}{0pt}{2pt}	
			Models  & Dataset & AbsRel& SqRel& RMS& RMS(${log}$) &  & $\delta_1$ & $\delta_2$ & $\delta_3$ \\ 
			\specialrule{0em}{0pt}{0pt}
			\midrule
			\specialrule{0em}{0pt}{2pt}	
			Monodepth~\cite{godard2017unsupervised}	&  K&  0.1480& 1.2550&	5.7320& 0.2250& & 0.808& 0.936& 0.973\\
			GeoNet~\cite{yin2018geonet}				&   K& 0.1550& 1.2960& 5.8570& 0.2230& & 0.793& 0.931& 0.973\\
			StructDepth~\cite{casser2019depth}		&   K& 0.1410& 1.0260& 5.2910& 0.2150& & 0.816& 0.945& 0.979\\
			BiCycDepth~\cite{wong2019bilateral}		&   K	& 0.1330 & 1.1260 & 5.5150 & 0.2310 & &0.826 & 0.934 & 0.969\\
			Monodepth2~\cite{godard2019digging}		&  K& 0.1150& 0.9030 & 4.8630& 0.1930& & 0.877& 0.959& 0.981\\		
			PackNet-SfM~\cite{guizilini20203d}		&   K& 0.1110& 0.7850 & 4.6010 & 0.1890& & 0.878 & 0.960 & 0.982\\
			SGDepth~\cite{klingner2020self}			&   K& 0.1170& 0.9070& 4.8440& 0.1960& & 0.875& 0.958& 0.980\\
			RMSFM6~\cite{zhou2021r}					&   K& 0.1120& 0.8060& 4.7040& 0.1910& & 0.878 & 0.960 & 0.981\\ 
			Mono-Former~\cite{bae2022monoformer}	&   K& 0.1080 & 0.8060& 4.5940 & 0.1840 & & 0.884& 0.963 & 0.983\\ 
			DRAFT~\cite{guizilini2022learning}		&   K& \textit{0.0970} & \textit{0.6470} & \textit{3.9910} & \textit{0.1690} & & \textit{0.899} & \textit{0.968} & \textit{0.984}\\  
			Ours 									&   K& \textbf{0.0950} & \textbf{0.6180} & \textbf{3.9400}	& \textbf{0.1680} & & \textbf{0.904} & \textbf{0.969} &\textbf{0.988}\\ \hline
			\specialrule{0em}{0pt}{0pt}	
			\bottomrule
		\end{tabular}
\end{table}

As experimental results in Table.\ref{tab:result_kitti}, the proposed method without pre-train outperforms other existing methods considerably.
The optimal metrics are denoted in bold, while the second-best results are indicated in italics. The proposed method achieves an AbsRel of 0.0950, a SqRel of 0.6180, an RMS of 3.940, and a RMS(${log}$) of 0.1680. Without pre-training, the proposed method reaches the highest accuracy across all metrics. Notably, the second-best depth estimation model, DRAFT~\cite{guizilini2022learning}, employs a large amount of ground truth optical flow for supervision, while our method is entirely self-supervised. Therefore, the proposed self-supervised method represents the optimal solution for depth estimation tasks.

\begin{figure}[h]
	\centering
	\includegraphics[width=\linewidth]{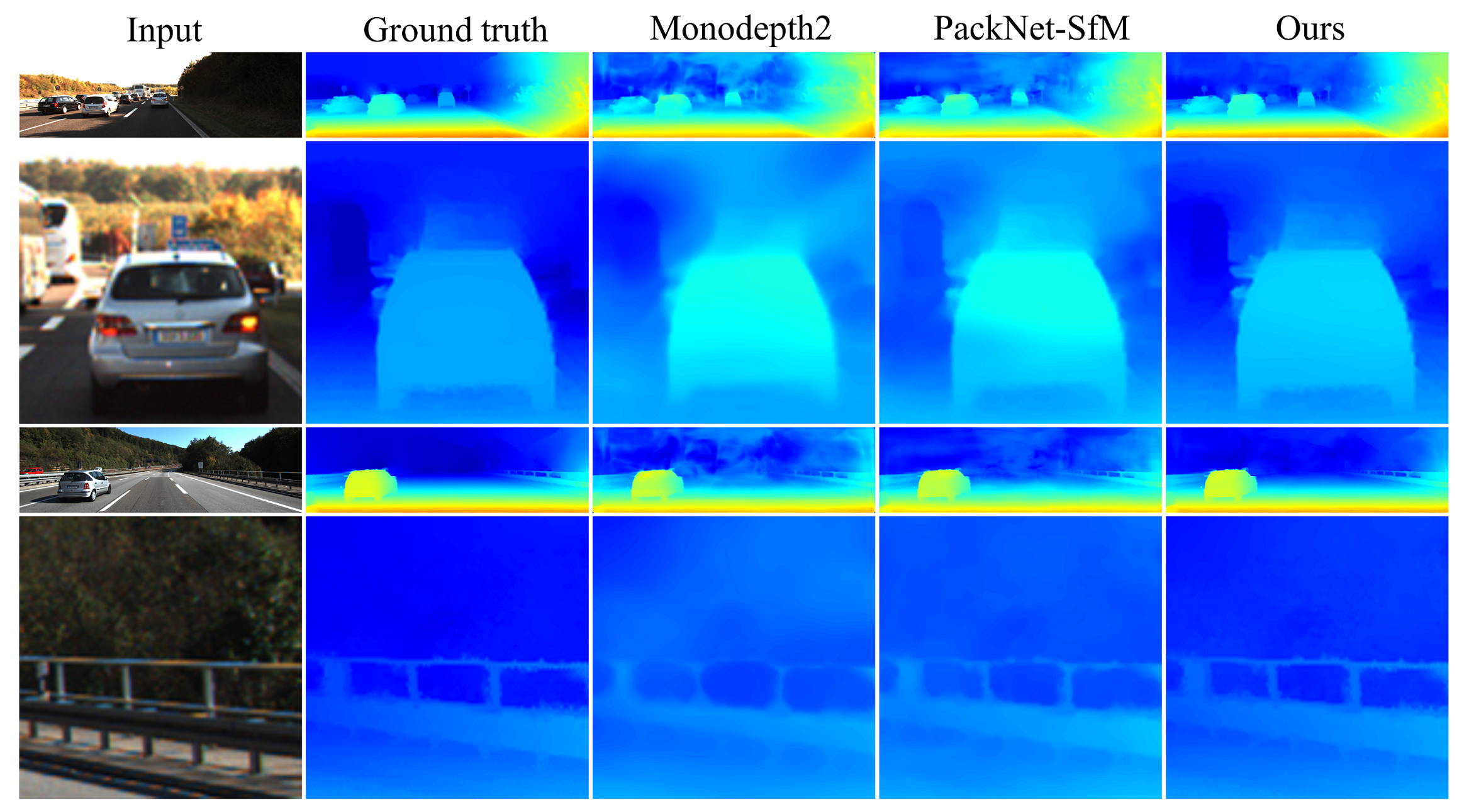}
	\caption{Visual results on KITTI pre-trained on CityScapes with inter-frame-supervised methods.}
	\label{fig:quantitive2}
\end{figure}

Following above evaluations, the experiments also compare our framework with existing methods on the KITTI dataset pre-trained Cityscapes. The visual results are presented in Fig.\ref{fig:quantitive2}, while the quantitative results are shown in Table.\ref{tab:result_kitti2}.

Fig.\ref{fig:quantitive2} displays the comparison between advanced methods and the proposed method with the pre-trained Cityscapes. All visual methods successfully reconstructed the depth maps of the lane scenes. Compare to other advanced methods, our motion segmentation-based joint optical flow and depth estimation method yields more accurate car edges in the upper image and neater road barriers in the lower image. Therefore, in the visual result comparison, the proposed method demonstrates higher accuracy on the depth estimation task.

\begin{table}[t]
	\caption{Quantitative depth results on Eigen split. Extensive depth estimators are trained on KITTI dataset with pre-trained CityScapes~\cite{cordts2016cityscapes} (K+CS).}
	\label{tab:result_kitti2}
		\begin{tabular}{l|c|cccc|ccccccc}
			\toprule
			\specialrule{0em}{0pt}{2pt}	
			Models  & Dataset & AbsRel& SqRel& RMS& RMS(${log}$) &  & $\delta_1$ & $\delta_2$ & $\delta_3$ \\ 
			\specialrule{0em}{0pt}{0pt}
			\midrule
			\specialrule{0em}{0pt}{1pt}	
			Monodepth~\cite{godard2017unsupervised}	&   K+CS& 0.1240& 1.0760& 5.3110& 0.2190& & 0.847& 0.942& 0.973\\
			GeoNet~\cite{yin2018geonet}				&   K+CS & 0.1530& 1.3280& 5.7370& 0.2320& & 0.802& 0.934& 0.972\\
			PackNet-SfM~\cite{guizilini20203d}		&   K+CS & 0.1080 & \textit{0.7270} & 4.4260 & 0.1840 & & 0.885 & \textit{0.963} & 0.982 \\
			BiCycDepth~\cite{wong2019bilateral}		&  K+CS & 0.1180& 0.9960 & 5.1340 & 0.2150 & & 0.849 & 0.945 & 0.975\\
			SGDepth~\cite{klingner2020self}			&  K+CS & 0.1170& 0.9070& 4.8440& 0.1960& & 0.875& 0.958& 0.980\\ 
			Mono-Former~\cite{bae2022monoformer}	&   K+CS & 0.1060 & 0.8390& 4.6270 & 0.1830 & & 0.889 & 0.962 & \textit{0.983}\\		
			SemanticGuide\cite{guizilini2020semantically}&	K+CS &	\textit{0.1000} &	0.7610 & \textit{4.2700} & \textit{0.1750} & &\textit{0.902} & \textit{0.965} &0.982\\
			Ours 									&   K+CS & \textbf{0.0940}& \textbf{0.6030}& \textbf{3.8920}& \textbf{0.1640}& & \textbf{0.905}& \textbf{0.973}&\textbf{0.989}\\ 
			\specialrule{0em}{0pt}{0pt}	
			\bottomrule
		\end{tabular}
\end{table}

As shown in Table.\ref{tab:result_kitti2}, 'K+CS' denotes the depth estimation model tested on the KITTI dataset and pretrained on the Cityscapes dataset. Our method exhibits an AbsRel of 0.0940, a SqRel of 0.6030, an RMS of 3.892, and an RMS(${log}$) of 0.1640. Similarly, all metrics for the pre-trained depth estimation model have achieved the highest accuracy.

\begin{table}[tbp]
	\centering
	\caption{Complexity comparison of existing depth estimation methods}
	\begin{tabular}{lrr}
		\toprule[1pt]
		Method 							& Model Size  &  Running Time \\ 
		\midrule[1pt]
		PackNet-SfM~\cite{guizilini20203d}        &     102.8M     		&	190.024ms\\
		Mono-Former~\cite{bae2022monoformer} 	  &    756.3M          		&	2302.958ms	\\ 
		Ours                        			  &          93.2M            	&	143.130ms  		\\      
		\bottomrule[1pt]
	\end{tabular}
	\label{tab:param}
\end{table}

To further analyze the model, we evaluate the model size and single-frame running run of existing depth estimation methods, with an input at the standard size of 352$\times$1216 in KITTI dataset.
As indicated in Table.\ref{tab:param}, the proposed method has the smallest parameter number, while the second-smallest size of PackNet-SfM~\cite{guizilini20203d} is 10.3\% larger.
Among depth estimators with similar accuracy, the complexity of the proposed method is much lower than other methods.

In summary, the qualitative and quantitative results of the depth estimation experiments demonstrate that the proposed joint depth and optical flow estimation method, based on optical flow segmentation, successfully reconstructs accurate depth maps of outdoor scenes with moving objects, surpassing most advanced methods in an efficacious way.

\subsection{Optical Flow Comparison with Existing Methods}

\begin{figure}[h]
	\centering
	\includegraphics[width=\linewidth]{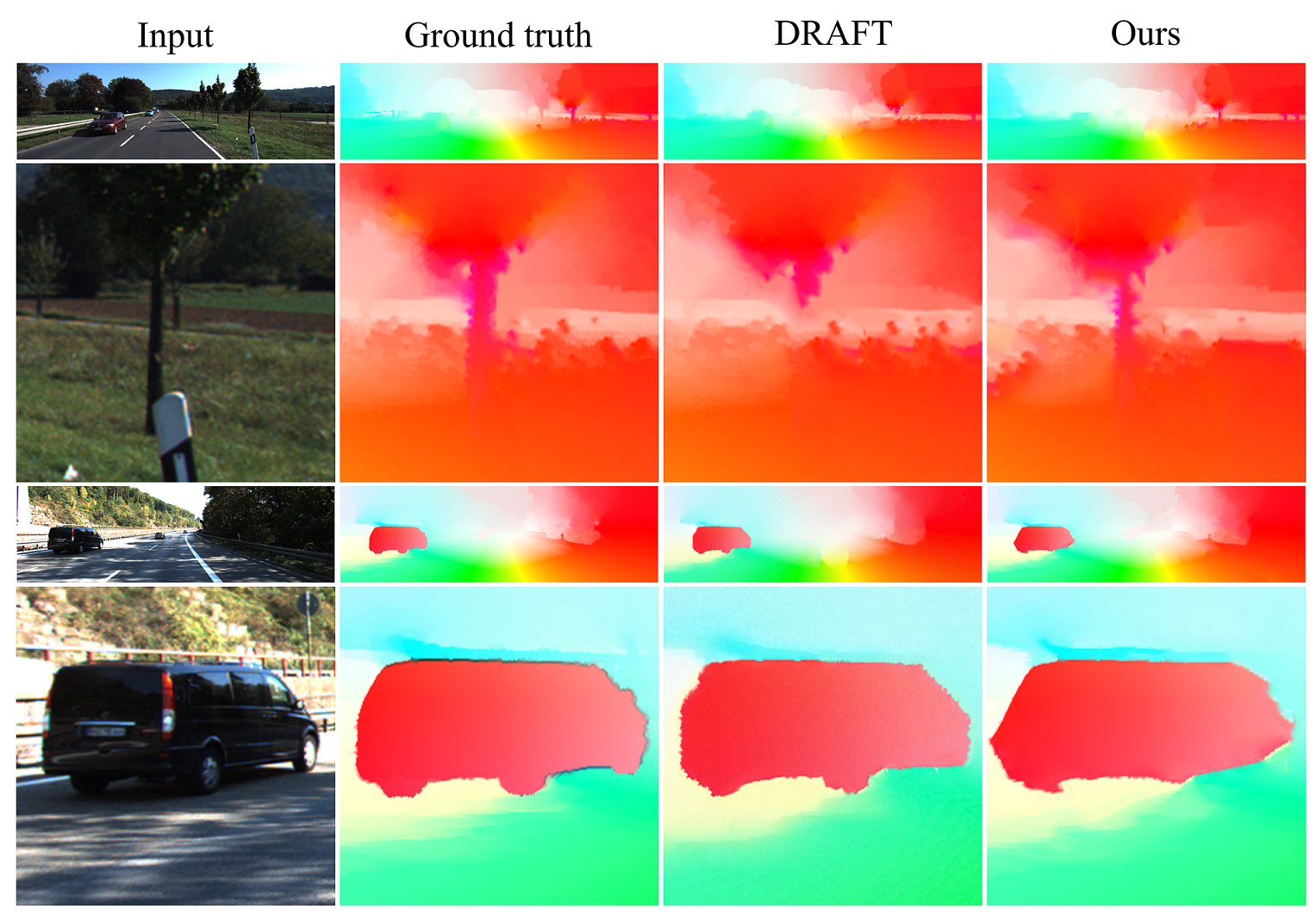}
	\caption{Visual optical flow results on KITTI pre-trained on CityScapes.}
	\label{fig:opticaldemo}
\end{figure}

In the joint task of depth and optical flow estimation, besides comparing the depth estimation experiment results, we conduct quantitative and qualitative comparisons of optical flow prediction results with existing methods. The visual results are provided in Fig\ref{fig:opticaldemo}, while the quantitative outcomes are shown in Table.\ref{tab:result_optical}.

As depicted in Fig.\ref{fig:opticaldemo}, the proposed method is visually compared with the DRAFT~\cite{guizilini2022learning}. From the optical flow estimation results, it can be observed that both the DRAFT method and the proposed method are visually accurate and reasonable, reconstructing the optical flow information of large areas with the same motion characteristics. Notably, compared to the previous best method, DRAFT, our approach offers more accurate reconstruction of moving object boundaries, such as the thin tree trunks in the upper image and the car contours in the lower image.

\begin{table}[t]
	\caption{Quantitative optical flow results on KITTI Flow dataset.}
	\label{tab:result_optical}
	\begin{tabular}{l|rr}
		\toprule
		\specialrule{0em}{0pt}{2pt}	
		Models    & EPE   & F1-all \\ 
		\specialrule{0em}{0pt}{0pt}
		\midrule
		\specialrule{0em}{0pt}{2pt}	
		HDD~\cite{yin2019hierarchical}	&	13.70& 24.00\\
		PWCNet~\cite{sun2018pwc}	  		& 	10.35 & 33.70  \\
		FlowNet2~\cite{ilg2017flownet}  	& 	10.10 & 29.90   \\
		DFNet~\cite{zou2018df}			&	8.98&26.00\\
		RAFT~\cite{teed2020raft}      	& 	5.04  & 17.40   \\
		TrianFlow~\cite{zhao2020towards} & 	3.60  & 18.05  \\
		DRAFT~\cite{guizilini2022learning}    & 2.55  & \textbf{14.81}   \\
		\specialrule{0em}{0pt}{0pt}
		\midrule
		\specialrule{0em}{0pt}{2pt}	
		Ours      &  \textbf{2.43} & 15.63   \\
		\specialrule{0em}{0pt}{0pt}	
		\bottomrule 
	\end{tabular}
\end{table}

As shown in Table.\ref{tab:result_optical}, the proposed method achieves the best optical flow accuracy in the EPE metric and second-best in the F1-all metric. Compared to the second-best optical flow estimation method, DRAFT, although our method's error increases by 5.53\% in the F1-all metric, it reduces the error by 4.70\% in the EPE metric. Consequently, our approach remains highly competitive in the optical flow estimation task.

Above experimental results demonstrate that the proposed method successfully reconstructs optical flow maps of outdoor scenes with various moving objects in the optical flow estimation task, outperforming most advanced methods.

\section{Conclusion}

In this work, we constrain the inter-frame-supervised depth and optical flow estimation, incorporating ego-motion segmentation to separate heterogeneous motion components.
Optical flow maps in a single motion direction can be equivalently decomposed into camera transformations and depths, allowing for independent depth and pose estimations in dynamic and static regions. 
Additionally, we treat ego-motion estimation in inter-frame supervision as a regression problem.
Further, optical flow synthesis derives from the inverse depth and ego-motion re-projections, aiming to penalize the errors between synthesis and preliminary estimates.
Resulting from the joint training with the two modules, optical flow and inter-frame-supervised depth module, extensive experiments confirm that the proposed framework yields the most advanced metrics on the KITTI depth dataset, both with and without pre-training on CityScapes.

\bibliography{cyc}

\end{document}